\documentclass[letterpaper, 10 pt, conference]{ieeeconf}  

\IEEEoverridecommandlockouts                              

\usepackage{cite}
\usepackage{amsmath,amssymb,amsfonts}
\usepackage{graphicx}
\usepackage{textcomp}
\usepackage{xcolor}
\usepackage{blindtext}
\usepackage{graphicx}
\usepackage{bm}
\usepackage{lipsum}
\usepackage{mwe}
\usepackage{babel}
\usepackage{cuted}
\usepackage{url}
\usepackage{hyperref}
\hypersetup{
    colorlinks=true,
    linkcolor=black,
    filecolor=magenta,    
    citecolor=black,
    urlcolor=blue,
    }
\usepackage{algorithm}
\usepackage{algorithmic}
\usepackage{multirow}
\usepackage{pifont}  
\usepackage{xcolor}   

\newcommand{\cmark}{\textcolor{green}{\ding{51}}} 
\newcommand{\xmark}{\textcolor{red}{\ding{55}}}   

\graphicspath{{images/}}
\allowdisplaybreaks

\title{\LARGE \bf
DPGP: A Hybrid 2D-3D Dual Path Potential Ghost Probe Zone Prediction Framework for Safe Autonomous Driving}

\author{Weiming Qu$^{1,\dagger}$, Jiawei Du$^{1,\dagger}$, Shenghai Yuan$^{3}$, Jia Wang$^{1}$, Yang Sun$^{4}$, Shengyi Liu$^{4}$,\\ 
Yuanhao Zhu$^{5}$, Jianfeng Yu$^{5}$, Song Cao$^{5}$, Rui Xia$^{5}$, Xiaoyu Tang$^{4}$, Xihong Wu$^{1,2}$, Dingsheng Luo$^{1,2,*}$%
\thanks{* Corresponding author: Dingsheng Luo (e-mail: dsluo@pku.edu.cn).}
\thanks{$\dagger$ Equal contribution.}%
\thanks{$^{1}$ National Key Laboratory of General Artificial Intelligence, Key Laboratory of Machine Perception (MoE), School of Intelligence Science and Technology, Peking University, Beijing 100871, China.}%
\thanks{$^{2}$ PKU-WUHAN Institute for Artificial Intelligence, Wuhan, China.}%
\thanks{$^{3}$ School of Electrical and Electronic Engineering, Nanyang Technological University, Singapore 639798, Singapore.}%
\thanks{$^{4}$ School of Instrument Science and Engineering, Southeast University, Nanjing 210018, China.}%
\thanks{$^{5}$ China Automotive Innovation Corporation, Nanjing, China.}%
}

\begin{document}


 \maketitle


\thispagestyle{plain}
\pagestyle{plain}

\begin{abstract}
Modern robots must coexist with humans in dense urban environments. A key challenge is the ghost probe problem, where pedestrians or objects unexpectedly rush into traffic paths. This issue affects both autonomous vehicles and human drivers. Existing works propose vehicle-to-everything (V2X) strategies and non-line-of-sight (NLOS) imaging for ghost probe zone detection. However, most require high computational power or specialized hardware, limiting real-world feasibility. Additionally, many methods do not explicitly address this issue. To tackle this, we propose DPGP, a hybrid 2D-3D fusion framework for ghost probe zone prediction using only a monocular camera during training and inference. With unsupervised depth prediction, we observe ghost probe zones align with depth discontinuities, but different depth representations offer varying robustness. To exploit this, we fuse multiple feature embeddings to improve prediction. To validate our approach, we created a 12K-image dataset annotated with ghost probe zones, carefully sourced and cross-checked for accuracy. Experimental results show our framework outperforms existing methods while remaining cost-effective. To our knowledge, this is the first work extending ghost probe zone prediction beyond vehicles, addressing diverse non-vehicle objects. We will open-source our code and dataset for community benefit. 
\end{abstract}

 \section{Introduction}
Modern robots must coexist with humans in dense urban environments \cite{zhou2022high} where unexpected pedestrian or object intrusions, known as ghost probe incidents, create challenges for both autonomous and human drivers. These incidents \cite{shrivastava2014global} occur when traffic participants suddenly appear from occluded areas, increasing the risk of accidents. Conventional perception methods, such as object detection and semantic segmentation, focus on identifying predefined objects but fail to detect blind zones where ghost probe incidents originate.

Many existing works \cite{odagiri2023monocular} are \textbf{biased toward} full-body or half-body detections \cite{karthi2021evolution,reis2023real} or semantic segmentation \cite{ji2024sgba,cao2024mopa}. However, these approaches \cite{yin2024outram} often fail to address the problem of sudden object appearances \cite{shimomura2024potential}, as illustrated in Fig. 1. To mitigate this issue, alternative strategies such as vehicle-to-everything (V2X) communication \cite{zhou2020evolutionary} and non-line-of-sight (NLOS) imaging \cite{yang2023av} have been proposed. While V2X relies on extensive sensor networks, its large-scale deployment is hindered by significant infrastructure demands. NLOS imaging reconstructs hidden objects using lasers or coherent light sources but requires \textbf{expensive hardware} and \textbf{specialized pipelines}, limiting its practical applicability \cite{clancy2024wireless}. Other approaches also explore laser-based \cite{cao2022computational, iseringhausen2020non} or coherent light-based \cite{chen2020learned,lei2019direct} methods to reconstruct occluded objects from reflections, yet their high costs and technical complexity remain barriers to adoption. Odagiri et al. \cite{odagiri2023monocular} addressed this challenge by extracting depth discontinuities from depth maps and back-projecting them into three-dimensional space. Similarly, Shimomura et al. \cite{shimomura2024potential} generated risk region labels using depth maps and segmentation masks, but their approach was \textbf{limited to vehicle-induced ghost probe zones}, overlooking other potential obstacles such as walls.

\begin{figure}
\centering
  \includegraphics[width=8cm]{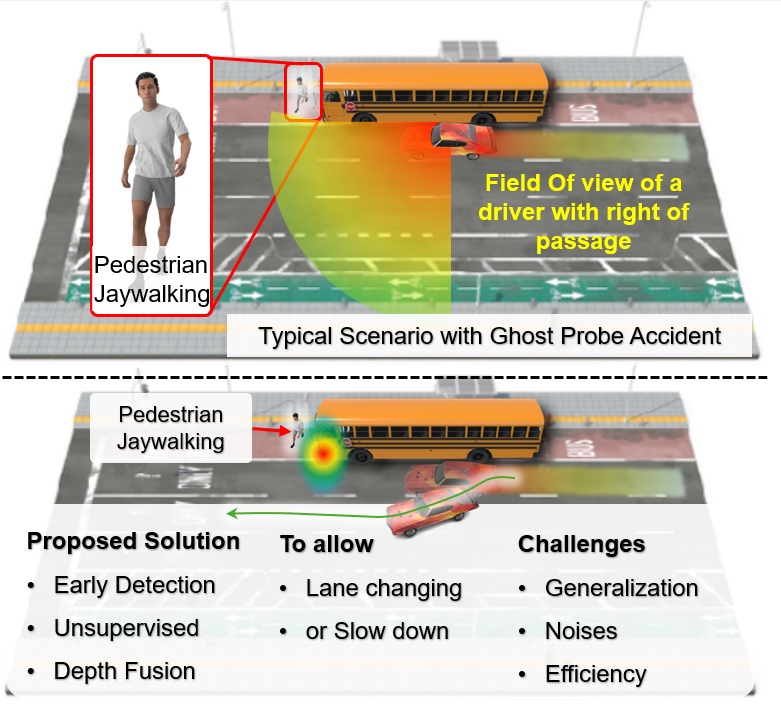}
      \vspace{-10pt}
\caption{Typical ghost probe accident scenario and proposed solution. 
The top section shows a pedestrian jaywalking, occluded by a bus, making them invisible to an approaching vehicle due to the driver's limited field of view. 
The bottom section illustrates the proposed solution, using early detection, unsupervised learning, and depth fusion to anticipate occluded pedestrians, enabling proactive actions like lane changing or slowing down. Key challenges include generalization, noise handling, and efficiency.}
  \label{fig:motivation}
      \vspace{-10pt}
\end{figure}

The \textbf{challenges} lie in predicting sudden object appearances in blind zones, which conventional perception fails to address. Existing solutions have limitations: vehicle-to-everything requires extensive infrastructure, non-line-of-sight imaging is costly and complex, and depth-based methods either focus only on vehicle-induced ghost probe zones or struggle with depth representation, as shown in Fig. 2. These constraints hinder reliable and practical ghost probe zone prediction.

To address these challenges, we propose DPGP, a hybrid 2D-3D fusion framework for ghost probe zone prediction. We first use Metric3Dv2 \cite{hu2024metric3d} for zero-shot monocular metric depth and surface normal estimation. Based on the hypothesis that blind spots align with depth discontinuities, we detect obstacle edges using depth gradient magnitude and generate 3D point clouds from the depth map. Our framework then extracts 2D image features via U-Net \cite{ronneberger2015u} and 3D point cloud features via PointNet++ \cite{qi2017pointnet}, integrating them through a cross-attention mechanism to enhance prediction accuracy. Unlike previous methods, DPGP operates without vehicle segmentation, enabling the detection of ghost probe zones from any occluding structure.
In summary, the main contributions of this paper are summarized as follows:

\begin{itemize}
    \item \textbf{Hybrid 2D-3D Feature Fusion:} We propose a novel framework integrating monocular depth estimation with point cloud processing, enabling ghost probe zone prediction beyond vehicle-induced occlusions with the affordable camera-only solution.

    \item \textbf{Cross-Attention Multi-Modal Fusion:} A cross-attention mechanism is introduced to fuse 2D visual features and 3D geometric features, improving robustness against depth estimation errors and enhancing spatial awareness.
    
    \item \textbf{Benchmarking and Ablation Studies:} We construct the CAIC-G dataset, conduct extensive comparisons with state-of-the-art methods, and perform ablation studies to analyze the impact of different feature modalities. Our source code and dataset will be made open-source upon acceptance  \url{https://github.com/ziyue-YING/DPGP}. 
\end{itemize}





\begin{figure}
\centering
  \includegraphics[width=1\linewidth]{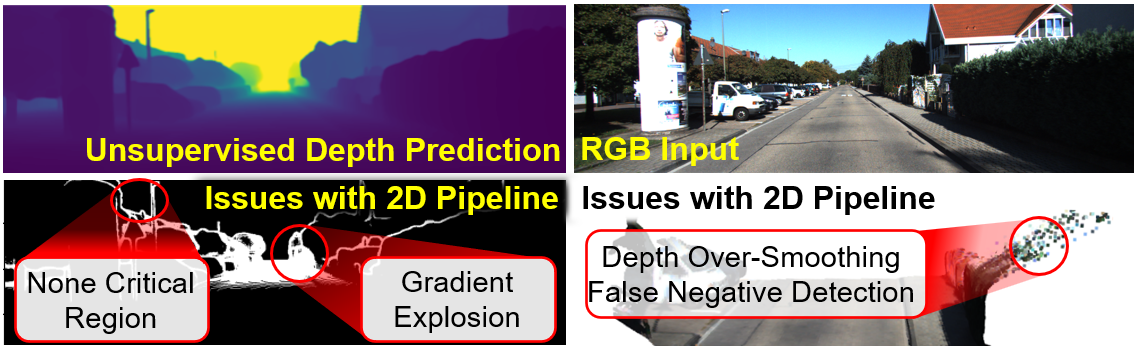}
      \vspace{-10pt}
\caption{Limitations of using only 2D images \cite{guo2018blind} or 3D point clouds \cite{qi2017pointnet++}: In the left image, the edge within the red box does not represent a potential ghost probe zone. In the right image, point cloud sparsity issue occurs within the red box.}
  \label{fig:issueswithexistingwork}
      \vspace{-10pt}
\end{figure}

\section{Related Works}
Extensive research has been conducted in blind spot leading to the development of various methods, which can be broadly classified into three categories: V2X-based methods, NLOS imaging-based methods, and 2D-3D Sensor-based methods.

V2X technology plays a crucial role in the advancement of safety systems for autonomous vehicles (AVs). Unlike traditional vehicular sensors, roadside sensing units (RSUs) are typically mounted at higher altitudes, which allows them to better address blind spots. Moreover, V2X enables real-time communication between vehicles, infrastructure, and other road users (RUs), offering a more comprehensive awareness of the surrounding environment. For instance, when an AV’s sensors fail to cover a blind spot, V2X can relay information about the location, speed, and orientation of nearby road users, helping the vehicle's perception system assess the potential risk of a collision. Maruta et al. \cite{maruta2021blind} proposed a blind-spot visualization framework that uses augmented reality (AR) glasses, leveraging millimeter-wave V2X communication for enhanced safety. Ni et al. \cite{ni2020v2x} introduced a method to avoid blind zone collisions between right-turning vehicles and pedestrians. Nourbakhshrezaei et al. \cite{nourbakhshrezaei2024novel} presented a novel ADAS using a computer vision-based background subtraction algorithm to enhance drivers' awareness of their surroundings, detecting objects in blind spots and providing real-time spatial and contextual information to improve decision-making and reduce collision risks. The system features a distributed architecture with cameras, processors, and local databases as RSUs, notifying drivers of nearby vulnerable road users, including pedestrians, cyclists, and AVs. However, while V2X-based methods show promise, they face significant challenges due to the need for widespread infrastructure deployment, including onboard and roadside sensors, as well as the management of complex communication protocols and large volumes of data, which make the large-scale implementation of V2X systems impractical in a short time.

NLOS imaging-based methods concentrate on recovering occluded objects, typically around corners or hidden by obstacles. Cao et al. [11] developed a physical NLOS imaging system along with a corresponding virtual setup to capture steady-state NLOS images under various lighting conditions. This system leverages both real and simulated NLOS images to train or fine-tune a multitask convolutional neural network architecture to perform simultaneous background illumination correction and NLOS object localization. However, the hardware is expensive and demands skilled technicians, leading to significant financial and technical barriers. In comparison to active NLOS imaging technologies, passive imaging techniques only require a commercial camera without an active light source, reducing the cost of necessary equipment \cite{seidel2020two}. However, it necessitates carefully controlled laboratory lighting, which is often impractical in real-world driving scenarios with numerous complex variables. To address this, Yan et al. \cite{yan2023ghost} designed a shadow signal discriminator to assess weak shadows cast by pedestrians, filtering out the influence of other complex lighting conditions. This approach can detect moving pedestrians in NLOS regions up to 20 meters away and issue warnings to the Advanced Driver Assistance System (ADAS) to maintain a safe distance. While this method demonstrates robust performance across various lighting environments and road conditions, it is not effective in handling rainy weather scenarios.

\begin{figure*}
\centering
  \includegraphics[width=1\linewidth]{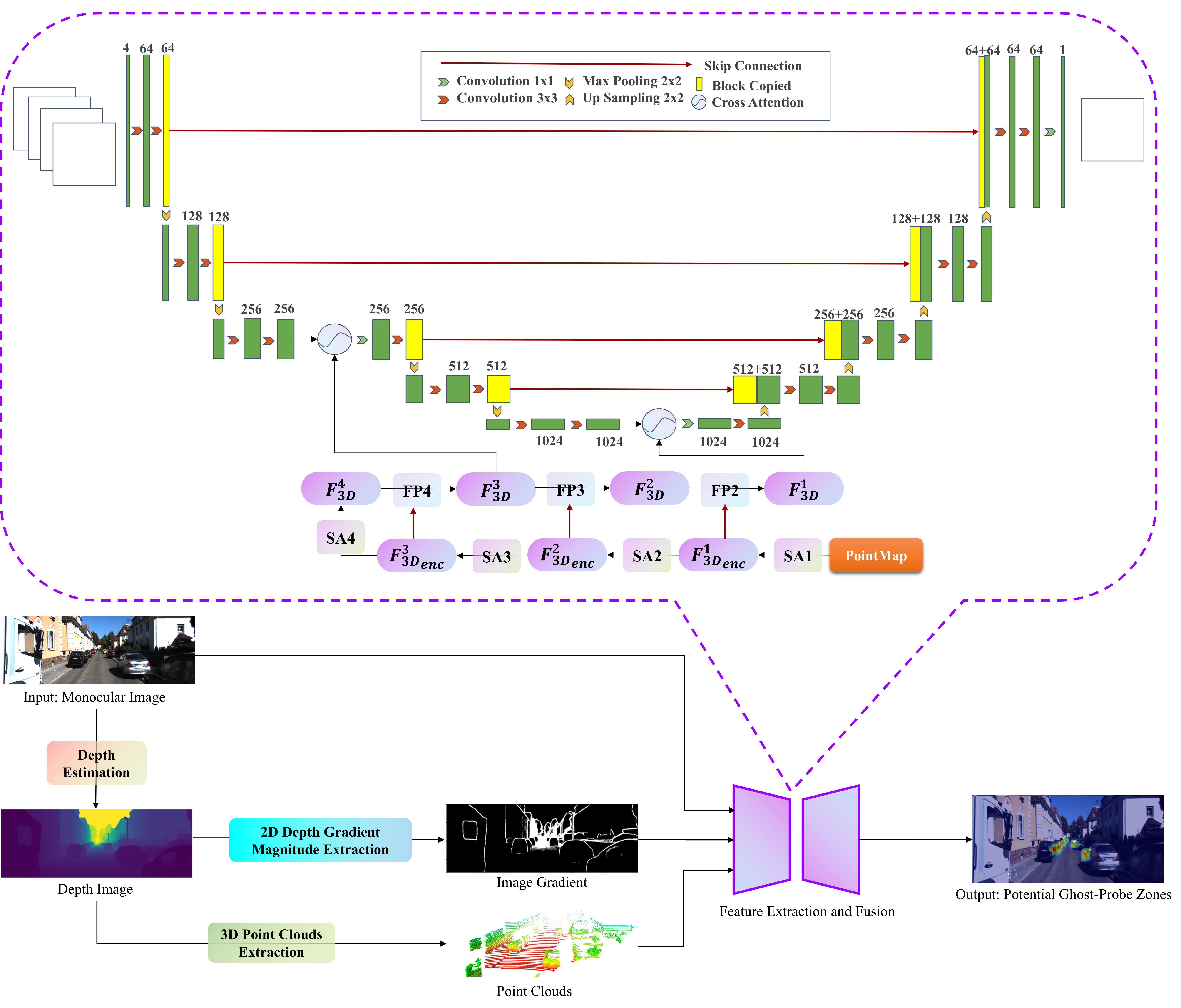}
      \vspace{-25pt}
\caption{Overview of the proposed 2D-3D feature fusion framework. The pipeline extracts depth, depth gradient, and 3D point clouds from a monocular image, processes them through a hierarchical encoder-decoder network with skip connections and cross-attention, and generates a fused representation for ghost probe detection.}
  \label{fig:flowchart}
      \vspace{-15pt}
\end{figure*}

To estimate blind spots in front of the ego vehicle, several methods based on Lidar and/or Camera are recently proposed. Fukuda et al. \cite{fukuda2022blindspotnet} proposed BlindSpotNet, which is able to automatically generate training labels by combining supervised depth estimation, semantic segmentation network, and SLAM. However, it requires substantial Lidar point clouds and large segmentation datasets for training. Sugiura et al. \cite{sugiura2019probable} proposed the use of generative adversarial network (GAN) to estimate potential occupancy grid maps (OGMs), accounting for uncertainty in blind spots. It combines Lidar and GPS/IMU sensors to train the network, aiming to improve accuracy in detecting blind spots. With advancements in computer vision and image-based analysis, alternative methods using monocular cameras have emerged as more cost-effective solutions. Guo et al. \cite{guo2018blind} demonstrated the ability of deep learning methods, based on monocular camera images, to accurately detect vehicles in the blind spot. The camera-based system extracts depth information from side-view camera images, enabling detection of obstacles in the blind spot area. Premachandra et al. \cite{premachandra2020detection} further improved upon this by developing a system using a 360° camera (3DVC), allowing for comprehensive data collection acrMethodsoss entire intersections. Earlier solutions \cite{zhou2022high} proposed a method that utilizes depth maps to detect blind spots and accelerate pedestrian detection by limiting the search area to regions around the blind spot. However, their method struggles in environments with numerous blind spots due to its computational cost and the limitation of detecting only the nearest blind spot.

\section{Methods}
In this section, we first introduce the general overview of DPGP. Next, we delve into the details of depth estimation. Finally, we elaborate on the crucial design steps of the feature extraction and fusion module.

\subsection{Framework Overview} \label{Framework_Overview}

The proposed DPGP framework is a hybrid 2D-3D dual path system designed to predict potential ghost probe zones. The workflow of the proposed method is summarized in Fig. 3. First, the depth estimation process extracts depth maps from the monocular image by Metric3Dv2 \cite{hu2024metric3d}. These depth maps are then processed to detect depth discontinuities, which are hypothesized to correspond to blind spots, by calculating the depth gradient magnitude through Scharr, allowing for the identification of potential ghost-probe zones. Simultaneously, 3D point clouds are generated from the depth map to capture more detailed spatial information. Subsequently, a novel feature extraction and fusion module is employed to extract both 2D image features and 3D point cloud features, and these features are integrated to predict potential ghost-probe zones. The key steps are detailed as follows.

\subsection{Depth Estimation} 
We adopt the state-of-the-art monocular depth estimation framework Metric3Dv2\cite{hu2024metric3d} for depth estimation, which enables zero-shot generalization across diverse camera configurations. This framework effectively addresses the challenge of recovering accurate metric depth from single images, even when focal lengths and sensor parameters vary. We first apply the Canonical Camera Space Transformation Module (CSTM) to project input images into a unified camera space, thereby eliminating metric ambiguity caused by differing intrinsic parameters. The backbone architecture uses a Vision Transformer (ViT-Large) pretrained on 142 million images, and is optimized with Random Proposal Normalization Loss (RPNL) to emphasize local geometric details, which is essential for identifying subtle depth variations in potential ghost-probe zones.

\subsection{Feature Extraction and Fusion Module} 
\subsubsection{2D Feature Extraction}
We employ a U-Net \cite{ronneberger2015u} architecture for pyramid-style feature extraction from 2D images. In this network, the contracting path progressively expands the receptive field through successive down-sampling operations to extract abstract features. On the other hand, the expansive path recovers spatial information by up-sampling step by step, and the high-resolution features from the encoder are fused via skip connections. The skip connection mechanism inherently establishes cross-scale feature associations, thus providing structural support for explicit Cross-Attention fusion.

In the contracting path, each stage consists of two repeated \(3 \times 3\) convolutional layers followed by a Rectified Linear Unit (ReLU) activation and a \(2 \times 2\) max-pooling layer. Given that the input to the \(l\)-th layer is denoted by \(X \in \mathbb{R}^{W \times H \times C}\), where \(H\) is the spatial height, \(W\) is the spatial width, and \(C\) is the number of channels, the transformations are as follows:

\begin{align}
    \tilde{X}^l &= A^l * X^l + b^l \in \mathbb{R}^{W \times H \times 2C} \quad  \\ 
    X^l_{\text{skip}} &= \text{ReLU}(A^l * \tilde{X}^l + b^l) \in \mathbb{R}^{W \times H \times 2C} \quad  \\ 
    X^{l+1}_{\text{t}} &= \text{MaxPool}(X^l_{\text{skip}}) \in \mathbb{R}^{\frac{W}{2} \times \frac{H}{2} \times 2C} \quad 
\end{align}

where \( A^l \) and \( b^l \) represent the weights and biases for the convolution operations at the \( l \)-th layer.

In the expansive path, we first perform up-sampling using a \(2 \times 2\) transpose convolution \((W_{\text{trans}})\), which doubles the feature map dimensions and reduces the number of channels by half. The resulting up-sampled feature map is then concatenated with the corresponding encoded features from the skip connection. Following this, two \(3 \times 3\) convolutional layers are applied, followed by a ReLU activation:

\begin{align}
    X^{l-1}_{\text{up}} &= W_{\text{trans}} \circ X^l \in \mathbb{R}^{2W \times 2H \times \frac{C}{2}}  \\ 
    X^{l-1}_{\text{con}} &= \text{concat}(X^{l-1}_{\text{up}}, X^{l-1}_{\text{skip}}) \in \mathbb{R}^{2W \times 2H \times C}   \\ 
    X^{l-1}_{\text{t}} &= \text{Conv}_{3 \times 3}(\text{Conv}_{3 \times 3}(X^{l-1}_{\text{con}})) \in \mathbb{R}^{2W \times 2H \times \frac{C}{2}} 
\end{align}

\subsubsection{3D Feature Extraction}
We utilize PointNet++ \cite{qi2017pointnet} to progressively extract multi-scale features from the 3D point cloud, denoted as \( \{ F^l_{3D} \}_{l=1}^{4} \). The higher-level features \( F^4_{3D} \) capture global semantic information at the object level, while the mid-level features \( F^2_{3D} \) retain local details at the part level.

In Encoder, we use 4 set abstraction structures to achieve multi-level down-sampling. At each level of the set abstraction
(SA), the point set is processed and abstracted to generate a smaller set of points, which results in point features of varying scales. Specifically, suppose the current point set \( P \in \mathbb{R}^{B \times N \times (d+D)} \), where \( B \) is the batch size, \( N \) is the number of points in the set, \( d \) represents the point's coordinate dimensions, and \( D \) represents the point features dimensions. The set abstraction process is as follows: First, we use Farthest Point Sampling (FPS) strategy to sample uniformly from the current point set \( P^l \in \mathbb{R}^{B \times N_l \times (d+D)} \) to get a smaller set \( P^{l+1} \in \mathbb{R}^{B \times N_{l+1} \times (d+D)} \). FPS addresses the issue of uneven point cloud density, ensuring that features from key regions are prioritized, which provides the structural basis for subsequent cross-modal alignment. For each key point \( p_i \in P^{l+1} \), we then search for \( K \) neighboring points to form a local neighborhood, denoted as \( L(p_i) \in \mathbb{R}^{B \times N_{l+1} \times K \times (d+D)} \). Finally, the features are aggregated using PointNet \cite{qi2017pointnet} to form the feature representation for each point in the set.

In Decoder, we perform up-sampling by feature propagation (FP), including interpolation, skip connections and PointNet, to progressively recover discriminative features. For the \( l \)-th layer feature \( F^l_{3D} \in \mathbb{R}^{B \times N_l \times D_l} \), we upsample it to \( N_{l-1} \) points:

\begin{align}
    F^{l-1}_{3D_{\text{dec}}} &= \sum_{i=1}^{K} \omega_i F^l_{3D} (p_i), \quad \omega_i = \frac{\frac{1}{\| x - p_i \|_2}}{\sum_{j=1}^{K} \frac{1}{\| x - p_j \|_2}} \quad 
\end{align}

where, \(\{ p_i \}_{i=1}^{K} \) represents the \( K \) nearest points to the target point \( x \) in \( F^l_{3D} \). After that, through skip connections, the up-sampled feature \( F^{l-1}_{3D_{\text{dec}}} \) is concatenated with the encoder feature \( F^{l-1}_{3D_{\text{enc}}} \) of the same scale as shown in Equation (2). 

This process effectively integrates multi-scale features and allows for the capture of both global and local information from the 3D point cloud.

\begin{align}
    F^{l-1}_{3D} &= \text{Conv}_{1 \times 1} \big( \text{concat}(F^{l-1}_{3D_{\text{dec}}}, F^{l-1}_{3D_{\text{enc}}}) \big) \\
    F^{l-1}_{3D} &\in \mathbb{R}^{B \times N_{l-1} \times (D_{l-1} + D_{l-1})} \quad 
\end{align}

\subsection{2D-3D Feature Fusion}
Cross-Attention is a powerful deep learning mechanism that establishes effective associations between data from different modalities. The core idea behind Cross-Attention is that the attention mechanism enables information from different sources to interact and complement each other within a specific context, thereby improving the quality of feature fusion. 2D features typically provide dense semantic information (such as texture and color), while 3D features usually offer geometric structural information (such as depth and shape). By fusing 2D and 3D features through Cross-Attention, the model can leverage the advantages of both modalities simultaneously.

In the proposed mechanism, we take the 2D image features \( F^{i}_{2D} \in \mathbb{R}^{H \times W \times C} \) as the Query, and the 3D point cloud features \( F^{i}_{3D} \in \mathbb{R}^{N \times D} \) as the Key and Value. The similarity between the Query and Key is computed dynamically to select the most relevant 3D features corresponding to the current image features. The process is described as follows:
    \vspace{-10pt}
\begin{align}
    Q &= \phi_{2D}(F_{2D}) W_Q, \hspace{0.25em} K = F_{3D} W_K,\hspace{0.25em}  V = F_{3D} W_V \quad 
\end{align}
\vspace{-10pt}
\begin{align}
    F_{\text{fuse}} &= \text{softmax} \left( \frac{Q}{\sqrt{\text{dim}}} \cdot K^T \right) \cdot V 
\end{align}

where \( \phi_{2D} \) represents the spatial flattening operation applied to the 2D image feature, \(\text{dim}\) represents the latent space dimension, \( W_Q \in \mathbb{R}^{C \times \text{dim}} \), \( W_K, W_V \in \mathbb{R}^{D \times \text{dim}} \) are learnable weight matrices.

\section{Experiments}

\subsection{Environmental Setup}

\textbf{Datasets Selection:} We collect a new dataset annotated with potential ghost-probe zones, termed CAIC-G. The dataset consists of over 500 images collected from Nanjing, China. The dataset annotation manner is labeled using the same technique as prior arts \cite{shimomura2024potential}. Additionally, we also evaluate the proposed method on the KITTI dataset. Specifically, we select 10 scenes from KITTI that contain a high number of potential ghost-probe zones, including scenes 00, 05, 06, 07, 08, 11, 13, 14, 15, and 18. For each selected scene, we annotate 50 representative images in the same manner as the CAIC-G dataset. To ensure fair comparisons, both our method and baseline methods are evaluated under identical experimental settings. By leveraging these two datasets, we aim to thoroughly evaluate the models’ prediction capability on various data distributions.

\textbf{Baselines Selections:} To the best of our knowledge, most existing works on ghost probe or blind spot detection \cite{zhou2022high, odagiri2023monocular, shimomura2024potential, maruta2021blind, ni2020v2x, nourbakhshrezaei2024novel, fukuda2022blindspotnet, premachandra2020detection} are not open-source, making replication difficult due to missing key parameters. We use available online implementations \cite{ren2016faster, tan2020efficientdet, item30, lv2024rt} for baseline comparisons. Additionally, we conduct an ablation study to demonstrate the contribution of each proposed module.

\textbf{Evaluation Metrics:} We adopt the standard detection evaluation metrics to measure the potential ghost probe zone prediction performance, including Recall (Detection Rate, DR), Precision (Detection Precision, DP), and F1-Score (F). Relevant formulas are defined as follows:
\begin{align}
    \text{Recall} &= \frac{TP}{TP + FN} \quad  \\
    \text{Precision} &= \frac{TP}{TP + FP} \quad  \\
    F1\text{-Score} &= \frac{2DR \cdot DP}{DR + DP} \quad 
\end{align}

where \( TP \) represents the number of true positive samples, \( FP \) represents the number of false positive samples, and \( FN \) represents the number of false negative samples.

\subsection{Implementation Details}
All our experiments were conducted on a single RTX 3090 GPU. During the training of the model, we employed the Adam optimizer with a learning rate set to 0.0001, and the training process spanned 300 epochs with batch size of 4. For both the KITTI and CAIC-G datasets, we allocated 80\% of the data for training purposes, while the remaining 20\% was reserved for validation. In our standard model configuration, both the RGB images and gradient maps were resized to a resolution of \(640 \times 640\) pixels.

\subsection{Quantitative Results}
The performance comparison of different models across different datasets is summarized in Table I. On the KITTI dataset, Faster R-CNN demonstrates a high detection rate. However, its relatively low precision reveals a major drawback of two-stage detectors, excessive candidate region generation in complex occlusion scenarios, leading to frequent false positives. YOLOv8 significantly improves precision by introducing a dynamic detection head, yet its anchor-based design constrains recall, reflecting the inherent trade-off between miss rate and false detection rate. RT-DETRv2, leveraging the end-to-end DETR framework, achieves a more balanced performance, validating the advantage of Transformer architectures in feature interaction. Our proposed method, by fusing 2D image feature and 3D point cloud information, not only maintains a high detection rate but also significantly enhances detection precision, achieving an F1-score improvement of over 4\% compared to the best-performing baseline (RT-DETRv2).

\begin{table}[t]
    \centering
    \caption{Performance Comparison of Different Methods.  \textbackslash denotes failure to predict.}
    \vspace{-10pt}
    \begin{tabular}{ccccc}
        \hline
        \hline
        \textbf{Dataset} & \textbf{Method} & \textbf{Recall} $\uparrow$ & \textbf{Precision} $\uparrow$ & \textbf{F1-Score} $\uparrow$ \\
        \hline
        \multirow{4}{*}{KITTI} 
        & Faster-RCNN & 0.9333 & 0.5060 & 0.6562 \\
        & EfficientDet & \textbackslash & \textbackslash & \textbackslash \\
        & YOLOv8 & 0.6778 & 0.9104 & 0.7771 \\
        & RT-DETRv2 & \textbf{0.9389} & 0.8756 & 0.9062 \\
        & Ours & 0.9150 & \textbf{0.9761} & \textbf{0.9446} \\
        \hline
        \multirow{4}{*}{CAIC-G} 
        & Faster-RCNN & 0.7619 & 0.5053 & 0.6076 \\
        & EfficientDet & \textbackslash & \textbackslash & \textbackslash \\
        & YOLOv8 & 0.7619 & \textbf{0.8571} & 0.8067 \\
        & RT-DETRv2 & 0.9206 & 0.7945 & 0.8529 \\
        & Ours & \textbf{0.9234} & 0.8408 & \textbf{0.8801} \\
        \hline
    \end{tabular}
    \label{table:performance_comparison}
       \vspace{-15pt}
\end{table}

\begin{figure*}
\centering
  \includegraphics[width=1\linewidth]{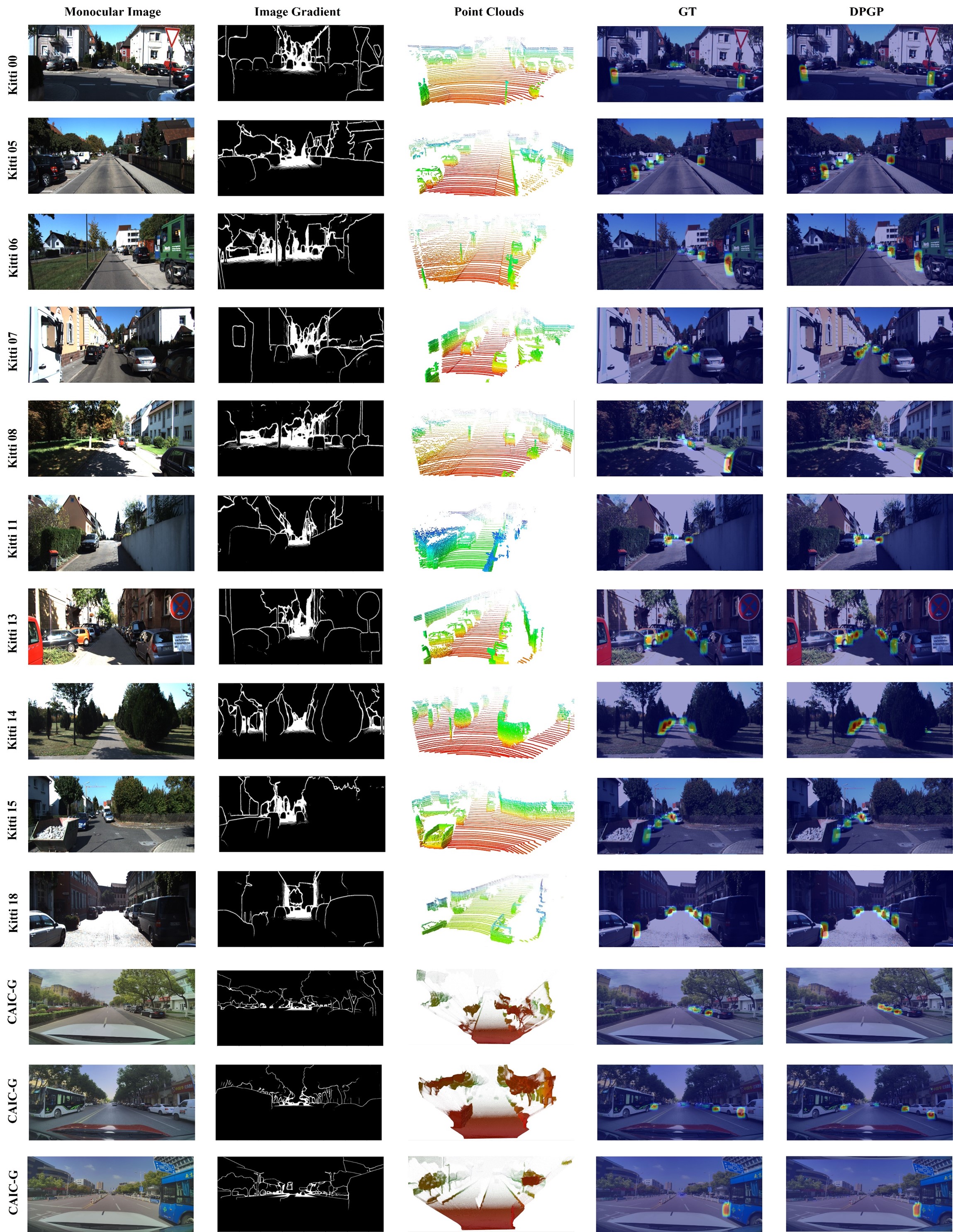}
\caption{Qualitative results of DPGP. We selected 1 sample from each scene in the KITTI dataset and 3 samples from the CAIC-G dataset for visualization}
  \label{fig:results}
\end{figure*}

The CAIC-G dataset results further highlight the model's generalization capability. Our DPGP maintains the highest detection rate in cross-domain testing. Although its precision is slightly lower than YOLOv8, it achieves the highest F1-score by maintaining a superior balance between recall and precision. In contrast, RT-DETRv2 exhibits a significant drop in precision on this dataset, indicating its sensitivity to domain shifts and scene variations. Additionally, EfficientDet fails to converge on both datasets, likely due to inherent limitations in its compound scaling approach.

Overall, the experimental results demonstrate that our 2D-3D feature fusion strategy effectively improves the model's ability to predict potential ghost probe zones. By maintaining detection sensitivity while significantly reducing false positives, our DPGP framework is particularly critical for autonomous driving applications, where minimizing false alarms is essential for reliable decision-making.

\subsection{Qualitative Results}
We present a qualitative analysis of DPGP to demonstrate its effectiveness in predicting potential ghost-probe zones. As shown in Fig. 4, our model is capable of accurately and comprehensively predicting the areas where ghost probes are likely to occur.

Notably, our model does not rely on any prior knowledge or segmentation of the vehicle's mask, which gives it a significant advantage. This allows our system to not only predict detection issues caused by vehicles but also to identify those caused by other environmental obstacles such as walls (KITTI 14), trees (KITTI 07), trees (KITTI 14), and off-path road borders (KITTI 15). This broadens the scope of ghost-probe detection, making it more robust and adaptable in dynamic environments. To the best of our knowledge, our work is the first to extend the prediction of ghost-probe zones beyond just vehicles, addressing scenarios where various non-vehicle objects may contribute to such issues. This capability demonstrates the versatility and robustness of our approach in real-world applications, particularly in complex environments containing mixed obstacle types, which is significant for safe autonomous driving.

Interestingly, in the KITTI 07 scene, the annotators did not label the area in front of the left white truck's front bumper. However, from an objective perspective, that region is indeed a blind spot and should be identified as a potential ghost probe zone. DPGP successfully detects this area, which demonstrates the strong generalization capability.

\subsection{Ablation Studies}
We conducted extensive ablation studies across both datasets to evaluate the individual contributions of various feature modalities in our DPGP framework. The results, as shown in Table II, systematically compare different model configurations. The experiments reveal that excluding any single feature modality results in noticeable performance degradation, further highlighting the complementary nature of each modality within the framework.

\begin{table}[t]
    \centering
    \caption{Ablation Studies in DPGP. The results demonstrate that the proposed solution effectively addresses the challenges of both 2D and 3D pipelines. Only by fusing both cases can we achieve significantly better and more accurate results.}
    \vspace{-10pt}
    \renewcommand{\arraystretch}{1.12}  
\renewcommand{\tabcolsep}{4pt}
    \begin{tabular}{ccccccc}
        \hline
        \hline
        \textbf{Dataset} & \textbf{RGB (2D)} & \textbf{IG (2D)} & \textbf{PCD (3D)} & \textbf{DR} $\uparrow$ & \textbf{DP} $\uparrow$ & \textbf{F} $\uparrow$ \\
        \hline
        \multirow{4}{*}{KITTI} 
        & \xmark & \xmark & \cmark & 0.2875 & 0.2896 & 0.2885 \\
        & \cmark & \cmark & \xmark & 0.8660 & 0.9434 & 0.9030 \\
        & \cmark & \xmark & \cmark & 0.8700 & 0.7464 & 0.8035 \\
        & \cmark & \cmark & \cmark & \textbf{0.9150} & \textbf{0.9761} & \textbf{0.9446} \\
        \hline
        \multirow{4}{*}{CAIC-G} 
        & \xmark & \xmark & \cmark & 0.4993 & 0.3689 & 0.4243 \\
        & \cmark & \cmark & \xmark & 0.8907 & 0.7452 & 0.8115 \\
        & \cmark & \xmark & \cmark & 0.8949 & 0.6210 & 0.7332 \\
        & \cmark & \cmark & \cmark & \textbf{0.9234} & \textbf{0.8408} & \textbf{0.8801} \\
        \hline
    \end{tabular}
    \label{table:ablation_studies}
        \vspace{-15pt}
\end{table}

Although the model that relied exclusively on 2D visual features (RGB+IG) shows a relatively high detection rate, its detection precision is quite low. This is due to the absence of 3D features, which leads to a higher false detection rate, as mentioned before in Fig.~2. Furthermore, when the model relied mainly on 3D point cloud features (IG+PCD), its performance was very poor, which directly aligns with the challenges outlined in Section I. These findings reinforce the necessity of integrating both 2D and 3D information, validating the effectiveness of our 2D-3D feature fusion strategy.

Another noteworthy observation from the experiments is that while incorporating depth gradient maps improved model performance, the enhancement was not as substantial as expected. One plausible explanation for this is that models without depth gradient maps as input still learned some depth gradient-related features. This observation indirectly supports our hypothesis that blind spots are associated with areas exhibiting depth discontinuities, as models without explicit gradient input still managed to capture relevant depth-related features through other modalities.

\section{Conclusion}
In this paper, we propose DPGP, a hybrid 2D-3D dual-path framework that enables accurate prediction of potential ghost-probe zones by fusing 2D image features and 3D point cloud representations. Unlike V2X or NLOS-based methods, which require expensive additional hardware, our approach relies solely on a monocular camera, making it a cost-effective solution. Experimental results demonstrate that DPGP significantly outperforms existing detectors on both the KITTI and CAIC-G datasets. The proposed method effectively maintains a high detection rate while significantly reducing false alarms, enhancing its reliability in real-world applications.

To the best of our knowledge, this is the first work to extend ghost-probe zone prediction beyond vehicle-induced blind spots, addressing scenarios where various non-vehicle objects, such as walls, gates, trees, or other static obstacles, may contribute to such risks.

For future work, we aim to explore adaptive driving strategies based on DPGP’s ghost-probe predictions to enhance safety in complex environments such as university campuses, senior living communities, and pedestrian-dense urban areas, where occlusions and unexpected pedestrian movements are more prevalent. This would further facilitate safe autonomous driving in highly dynamic, unstructured, and constrained spaces.

\section*{Acknowledgment}
The work is supported in part by the National Natural Science Foundation of China (No. 62176004, No. U1711327), Intelligent Robotics and Autonomous Vehicle Lab (RAV), Wuhan East Lake High-Tech Development Zone National Comprehensive Experimental Base of Governance of Intelligent Society, and High-performance Computing Platform of Peking University.

\bibliographystyle{IEEEtran}
\bibliography{IEEEfull}

\end{document}